# Statistical linear estimation with penalized estimators: an application to reinforcement learning

**Bernardo Ávila Pires**                                                                      BPIRES@UALBERTA.CA
Department of Computing Science, University of Alberta, Edmonton, AB T6G2E8 CANADA

**Csaba Szepesvári**                                                                      SZEPESVA@CS.UALBERTA.CA
Department of Computing Science, University of Alberta, Edmonton, AB T6G2E8 CANADA

## Abstract

Motivated by value function estimation in reinforcement learning, we study statistical linear inverse problems, *i.e.*, problems where the coefficients of a linear system to be solved are observed in noise. We consider penalized estimators, where performance is evaluated using a matrix-weighted two-norm of the defect of the estimator measured with respect to the true, unknown coefficients. Two objective functions are considered depending whether the error of the defect measured with respect to the noisy coefficients is squared or unsquared. We propose simple, yet novel and theoretically well-founded data-dependent choices for the regularization parameters for both cases that avoid data-splitting. A distinguishing feature of our analysis is that we derive deterministic error bounds in terms of the error of the coefficients, thus allowing the complete separation of the analysis of the stochastic properties of these errors. We show that our results lead to new insights and bounds for linear value function estimation in reinforcement learning.

## 1. Introduction

Let $A$ be a real-valued $m \times d$ matrix, $b$ be a real-valued $m$-dimensional vector, $M$ be an $m \times m$ positive semi-definite matrix, and consider the loss function $\mathcal{L}_M : \mathbb{R}^d \to \mathbb{R}$ defined by

$$\mathcal{L}_M(\theta) \doteq \|A\theta - b\|_M \,,$$

where $\|\cdot\|_M$ denotes the $M$ matrix-weighted two-norm. We consider the problem of finding a minimizer of this loss when instead of $A, b$, one has access only to their respective "noisy" versions, $\hat{A}, \hat{b}$. We call this problem a *statistical linear inverse* problem.

Our main motivation to study this problem is to better understand the so-called *least-squares approach to value function estimation* in reinforcement learning, whose goal is to estimate the value function that corresponds to a Markov reward process.[1] The least-squares approach originates from the work of Bradtke and Barto (1996), who proposed to find the parameter-vector $\hat{\theta}$ of a linear-in-the-parameters value function by solving $\hat{A}\theta = \hat{b}$ where the "noisy" matrix-vector pair, $(\hat{A}, \hat{b})$, is computed based on a finite sample. They have proven the almost sure convergence of $\hat{\theta}$ to $\theta^*$, the solution of $A\theta = b$, under appropriate conditions on the sample as the sample-size converges to infinity. In particular, they assumed that the sample is generated from either an absorbing or an ergodic Markov chain. More recently, several studies appeared where the *finite-sample performance* of LSTD-like procedures were investigated (see, *e.g.*, (Antos et al., 2008; Ghavamzadeh et al., 2010; Lazaric et al., 2010; Ghavamzadeh et al., 2011)). The nonparametric variant has also received some attention (Farahmand et al., 2009; Maillard, 2011).

One of the difficulties in the analysis of these procedures is that in these problems the sample is correlated, so the standard techniques of supervised learning that assume independence cannot be used. The approach followed by the above-mentioned papers is to extend the existing techniques on an individual basis to deal with correlated samples. However, this might be quite laborious, even only considering the relatively easier case of regression[2] (*e.g.*, Farahmand and Szepesvári 2011). Thus, a more appealing approach might be to first derive error bounds as a function of the errors $\hat{A} - A$, $\hat{b} - b$. The advantage of this approach is that it allows one to decouple the technical issue of studying the concentration of the errors $\hat{A} - A$, $\hat{b} - b$ from the error (or stability) analysis of the estimation procedures. This is the approach that we advocate and follow in this paper. Consequently, our results will always be applicable when one can prove the concentration of the errors $\hat{A} - A$, $\hat{b} - b$, leading to an overall elegant, modular approach to deriving finite-sample bounds. In some way, our approach parallels the recent trend in learning theory where sharp finite-sample bounds are obtained by first proving deterministic "regret bounds" (*e.g.*, Cesa-Bianchi et al., 2004).

---

[1] For background on this problem the reader may consult, *e.g.*, the books by Bertsekas and Tsitsiklis (1996); Sutton and Barto (1998); Szepesvári (2010).

A second unique feature of our approach is that we derive our results in the above-introduced framework of general statistical linear inverse problems. This allows us to concentrate on the high-level structure of the problem and yields cleaner proofs and results. Furthermore, we think that the problem of linear estimation is interesting on its own due to its mathematical elegance and its applicability beyond value function estimation (a number of specific linear inverse problems, ranging from computer tomography to time series analysis, are discussed in the books by Kirsch (2011) and Alquier et al. (2011)).

We will also place special emphasis in statistical linear inverse problems whose underlying system is *inconsistent* (*i.e.*, when there is no solution to $A\theta = b$). In value function estimation, such inconsistency may arise in the so-called *off-policy* version of the problem. Understanding the inconsistent case is important because results that apply to it may shed light on issues arising when learning in badly conditioned systems.

### 1.1. Goals

In this paper, our goal will be to derive *exact, uniform, fast, high-probability oracle inequalities* for the estimation procedures we study. That is, our goal is to prove that for our choice of an estimator $\hat{\theta}$, for any $0 < \delta < 1$, with probability $1 - \delta$,

$$\mathcal{L}_M(\hat{\theta}) \leq \inf_{\theta} \left\{ \mathcal{L}_M(\theta) + c_{\hat{A},\hat{b}}(\theta, \delta) \right\}, \quad (1)$$

where, for fixed values of $\theta, \delta$,

$$c_{\hat{A},\hat{b}}(\theta, \delta) = O(\max(\|\hat{A} - A\|, \|\hat{b} - b\|)) \quad (2)$$

---

[2]Regression is a special case of value function estimation (Szepesvári, 2010).

for some appropriate norm $\|\cdot\|$. The above is called an oracle inequality since the performance of $\hat{\theta}$ (as measured with the loss) is compared to that of an "oracle" that has access to the true loss function. The term $c_{\hat{A},\hat{b}}(\theta, \delta)$ expresses the "regret" permitted due to the lack of knowledge of the true loss function. The scaling of this term with $\theta$ (or a norm of it) and $\delta$ will also be of interest.

Let us now explain the special attributes of the above inequality. We call the "rate" in the above inequality "fast" when (2) holds. Such a "fast rate" is possible in simple settings (*e.g.*, when $d = 1$, $A = \hat{A} = 1$), hence it is natural to ask whether such rates are still possible in more general settings. The oracle inequality above is called *exact* because the leading constant (the constant multiplying $\mathcal{L}_M(\theta)$) equals to 1. When $L^* = \inf_\theta \mathcal{L}_M(\theta)$ is positive (implying that the system is *inconsistent*), then only a leading constant of one can guarantee the convergence of the loss to the minimal loss, *i.e.*, the consistency *of the estimator*. We call the above inequality *uniform* because it holds for any value of $\delta$. This should be contrasted with inequalities where the range of $\delta$ is lower-bounded and/or the estimator uses its value as input, which may be useful in some cases but falls short of fully characterizing the tail behavior of the loss of the resulting estimator. With some abuse of terminology, an inequality of the above form that holds for all small values of $\delta$ shall be also called uniform. Uniform bounds seem to be harder to prove than their non-uniform counterparts, and we do not know of any uniform, high-probability exact oracle inequality with fast rates, not even in the case of linear regression. Unfortunately, we were also unable to derive such results.

When deriving the estimators, we shall see that a major challenge is to control the magnitude of $\hat{\theta}$. Indeed, it follows from our objective function that the size of $A\hat{\theta}$ must be controlled, and when $A$ is unknown the magnitude of $\hat{\theta}$ must be controlled. This might be difficult when following a naive approach of solving $\hat{A}\theta = \hat{b}$ to get $\hat{\theta}$, *e.g.*, when $\hat{A}$ is singular, or near-singular (as might be the case frequently in practice). To cope with this issue, in this paper we study procedures built around penalized estimators where a penalty $\text{Pen}(\theta)$ is combined with the empirical loss $\hat{\mathcal{L}}_M(\theta) = \|\hat{A}\theta - \hat{b}\|_M$. The penalty is assumed to be some norm of $\theta$. We study two procedures. In the first one, the loss is combined directly with the penalty, in an additive way to get the objective function $\hat{\mathcal{L}}_M(\theta) + \lambda\|\theta\|$, while in the second one the square of the empirical loss is combined with the penalty: $\hat{\mathcal{L}}_M^2(\theta) + \lambda\|\theta\|$. Note that both objective functions are convex. We note in passing that the second objective function when $\|\theta\|$ is the $\ell^1$-norm

gives a Lasso-like procedure, but we postpone further discussion of these choices to later sections of the paper.

In the case of both objective functions the main issue becomes selecting the regularization coefficient $\lambda > 0$. In this paper we give novel procedures to this end and show that these procedures have advantageous properties: we are able to derive oracle inequalities with fast rates for our procedures, although the inequalities will be either exact or uniform (but not both). To the best of our knowledge our general approach, our procedures, analytic tools and results are novel.

The organization of the paper is as follows: in the next section, to motivate the general framework, we briefly describe value function estimation and how it can be put into our general framework. This is followed by a brief section that gives some necessary definitions. Section 3 contains our main results for the two approaches mentioned above. Section 4 discusses the results in the context of value function estimation. The paper is concluded and future work is discussed in Section 5.

## 2. Value-estimation in Markov Reward Processes

The purpose of this section is to show how our results can be applied in the context of value-estimation in Markov Reward Processes. Consider a *Markov Reward Process (MRP)* $(X_0, R_1, X_1, R_2, \ldots)$ over a (topological) state space $\mathcal{X}$. By this we mean that $(X_0, R_1, X_1, R_2, \ldots)$ is a stochastic process, $(X_t, R_{t+1}) \in \mathcal{X} \times \mathbb{R}$ for $t \geq 0$ and given the history $\mathcal{H}_t = (X_0, R_1, X_1, R_2, \ldots, X_t)$ up to time $t$, the distribution of *state* $X_{t+1}$ is completely determined by $X_t$, while the distribution of the *reward* $R_{t+1}$ is completely determined by $X_t$ and $X_{t+1}$ given the history $\mathcal{H}_{t+1}$. Denote by $P_M$ the distribution of $(R_{t+1}, X_{t+1})$ given $X_t$. We shall call $P_M$ a *transition kernel*. Assume that support of the distribution of $X_0$ covers the whole state space $\mathcal{X}$. Define the *value of a state* $x \in \mathcal{X}$ by $V(x) = \mathbb{E}\left[\sum_{t=0}^{\infty} \gamma^t R_{t+1} | X_0 = x\right]$, where $0 < \gamma < 1$ is the so-called *discount factor*. One central problem in reinforcement learning is to estimate the *value function* $V$ given the trajectory $(X_0, R_1, X_1, R_2, \ldots)$ (Sutton and Barto, 1998). One popular method is to exploit that the value function is the unique solution to the so-called *Bellman equation*, which takes the form $TW - W = 0$, where $W : \mathcal{X} \to \mathbb{R}$ and $T : \mathbb{R}^{\mathcal{X}} \to \mathbb{R}^{X}$ is the so-called *Bellman operator* defined using $(TW)(x) = \mathbb{E}\left[R_{t+1} + \gamma W(X_{t+1}) | X_t = x\right]$. Note that $T$ is affine linear.

Given a finite sample $(X_0, R_1, X_1, R_2, \ldots, X_{n+1})$, the LSTD algorithm of Bradtke and Barto (1996) finds an approximate solution to the Bellman equation by solving the linear system

$$\sum_{t=1}^{n}(R_{t+1} + \gamma W_\theta(X_{t+1}) - W_\theta(X_t))\phi(X_t) = 0 \quad (3)$$

in $\theta \in \mathbb{R}^d$. Here $\phi = (\phi_1, \ldots, \phi_d)^\top$ is a vector of $d$ basis functions, $\phi_i : \mathcal{X} \to \mathbb{R}$, $1 \leq i \leq d$, and $W_\theta : \mathcal{X} \to \mathbb{R}$ is defined using $W_\theta(x) = \langle \theta, \phi(x) \rangle$. Denoting by $\hat\theta$ the solution to (3), $W_{\hat\theta}$ is the approximate value function computed by LSTD. This method can be derived as an instrumental variable method to find an approximate fixed point of $T$ (Bradtke and Barto, 1996) or as a Bubnov-Galerkin method (Yu and Bertsekas, 2010). In any case, the method can be viewed as solving a "noisy" version of the linear system

$$A\theta = b. \quad (4)$$

Here, $A = \mathbb{E}\left[(\phi(X_t^{\text{st}}) - \gamma\phi(X_{t+1}^{\text{st}}))\phi(X_t^{\text{st}})^\top\right]$ and $b = \mathbb{E}\left[\phi(X_t^{\text{st}}) R_{t+1}^{\text{st}}\right]$, where $(X_0^{\text{st}}, R_1^{\text{st}} X_1^{\text{st}}, R_2^{\text{st}}, \ldots)$ is a steady-state MRP with transition kernel $P_M$.[3] The linear system (4) can be shown to be consistent (Bertsekas and Tsitsiklis, 1996).[4] Note that (3) can also be written in the compact form $\hat{A}\theta = \hat{b}$, where $\hat{A} = 1/n \sum_{t=1}^{n}(\phi(X_t) - \gamma\phi(X_{t+1}))\phi(X_t)^\top$ and $\hat{b} = 1/n \sum_{t=1}^{n} R_{t+1}\phi(X_t)$. By thinking of $\hat{A}, \hat{b}$ as "noisy" versions of $A$, $b$ and observing that for any $M \succ 0$ solutions to (4) coincide with the minimizers of $\mathcal{L}_M(\theta) = \|A\theta - b\|_M$ we see that the least-squares approach to value function estimation can be cast as an instance of statistical linear inverse problems. When $M = C^{-1}$, $C = \mathbb{E}\left[\phi(X_t)\phi(X_t)^\top\right]$, $\mathcal{L}_M(\cdot)$ becomes identical to the so-called *projected Bellman error loss* which can also be written as $\mathcal{L}_M(\theta) = \|\Pi_{\phi,\mu}(TW_\theta - W_\theta)\|_{\mu,2}$, where $\mu$ is the steady-state distribution underlying $P_M$, $\|\cdot\|_{\mu,2}$ is the weighted $L^2(\mu)$-norm over $\mathcal{X}$ and $\Pi : L^2(\mathcal{X}, \mu) \to L^2(\mathcal{X}, \mu)$ is the projection on the linear space spanned by $\phi$ with respect to the $\|\cdot\|_{\mu,2}$-norm (Antos et al., 2008).

Note that under mild technical assumptions (to be discussed later) one can show that $(\hat{A}_n, \hat{b}_n) = (\hat{A}, \hat{b})$ gets concentrated around $(A, b)$ at the usual parametric rate as the sample size $n$ diverges. Thus, we can indeed view $\hat{A}, \hat{b}$ as "noisy" approximations to $(A, b)$.

---

[3] The MRP is said to be in a steady-state if the distribution of $X_t$ is independent of $t$.

[4] For a discussion of how well $W_{\theta^*}$ approximates $V$ the reader is directed to consult the paper by Scherrer (2010) and the references therein. In this paper, we do not discuss this interesting problem but accept (4) as our starting point.

One variation of this problem, the so-called *off-policy* problem, gives further motivation to recast the problem in terms of a loss function $\mathcal{L}_M(\cdot)$ to be minimized. In the off-policy problem the data comes in the form of triplets, $((X_0, \tilde{R}_1, \tilde{X}_1), (X_1, \tilde{R}_2, \tilde{X}_2), \ldots)$, where the distribution of $(\tilde{R}_{t+1}, \tilde{X}_{t+1})$ is again independent of $\mathcal{H}_t = ((X_0, \tilde{R}_1, \tilde{X}_1), (X_1, \tilde{R}_2, \tilde{X}_2), \ldots, (X_{t-1}, \tilde{R}_t, \tilde{X}_t))$ given $X_t$ and is equal to the transition kernel $P_M$. Further, it is assumed that $(X_t)_{t \geq 0}$ is a Markov process. The previous setting (also called the on-policy case) is replicated when $\tilde{X}_t = X_t$, thus this new setting is more general than the previous one. The straightforward generalization of the least-squares approach is to define $A = \mathbb{E}\left[(\phi(X_t^{\text{st}}) - \gamma\phi(\tilde{X}_{t+1}^{\text{st}}))\phi(X_t^{\text{st}})^\top\right]$ and $b = \mathbb{E}\left[\tilde{R}_{t+1}^{\text{st}} \phi(X_t^{\text{st}})\right]$ for the "steady-state" process $(X_t^{\text{st}}, \tilde{R}_{t+1}^{\text{st}}, \tilde{X}_{t+1}^{\text{st}})_{t \geq 0}$. In this case, the linear system $A\theta = b$ is not necessarily consistent but one can still aim for minimizing (for example) the projected Bellman error. Using $\hat{A} = 1/n \sum_{t=1}^n (\phi(X_t) - \gamma\phi(\tilde{X}_{t+1}))\phi(X_t)^\top$ and $\hat{b} = 1/n \sum_{t=1}^n \tilde{R}_{t+1}\phi(X_t)$ we can again cast the problem as a statistical linear inverse problem.

## 3. Results

In this section we give our main results for statistical linear inverse problems. We start with a few definitions. For real numbers $a, b$, we use $a \vee b$ to denote $\max(a, b)$. The operator norm of a matrix $S$ with respect to the Euclidean norm $\|\cdot\|_2$ is known to satisfy $\|S\|_2 = \nu_{\max}(S)$. In what follows, we fix a vector norm $\|\cdot\|$. Define the errors of $\hat{A}$ and $\hat{b}$ with the following respective equations: let

$$\Delta_A \doteq \|M^{\frac{1}{2}}(A - \hat{A})\|_{2,*}, \quad \Delta_b \doteq \|M^{\frac{1}{2}}(b - \hat{b})\|_2, \quad (5)$$

where $\|X\|_{2,*}$ denotes the operator norm of matrix $X$ with respect to the norms $\|\cdot\|_2$ and $\|\cdot\|$, meaning that $\|X\|_{2,*} = \sup_{v \neq 0} \|Xv\|_2 / \|v\|$.

Although our main results are oracle inequalities, it will also be interesting to name a minimizer of $\mathcal{L}_M(\theta)$ to explain the structure of some bounds. For this, we introduce $\theta^* \in \mathbb{R}^d$ as a vector such that $\theta^* \in \arg\min_{\theta \in \mathbb{R}^d} \mathcal{L}_M(\theta)$ where if multiple minimizers exist we choose one with the minimal norm $\|\cdot\|$. [5]

In general, $\Delta_A, \Delta_b$ are unknown. As it will turn out, in order to properly tune the penalized estimation methods we consider, we need at least upper bounds on these quantities (in particular, on $\Delta_A$). To stay independent of sampling assumptions, we assume that

---

[5] Since our loss function is convex one can always find at least one minimizer.

suitable high-probability bounds on $\Delta_A$ and $\Delta_b$ are available:

**Assumption 3.1.** *There exist known scaling constants $s_A, s_b > 0$ and known "tail" functions $z_{A,\delta}, z_{b,\delta}$, $\delta \in (0, 1]$ s.t. for any $0 < \delta < 1$, the following hold simultaneously with probability (w.p.) at least $1 - \delta$:*

$$\Delta_A \leq s_A z_{A,\delta}, \qquad \Delta_b \leq s_b z_{b,\delta}.$$

*To fix the scales of these bounds, we restrict $z_{A,\delta}, z_{b,\delta}$ so that $z_{A,\frac{1}{e}} = z_{b,\frac{1}{e}} = 1$, where $e$ is the base of natural logarithm.*

The reason to have two terms on the right-hand side in the above inequalities as opposed to having a single term only is because we wish to separate the terms attributable to $\delta$ and the sample size. The intended meaning of $s_a$ (and $s_b$) is to capture how the errors behave as a function of the sample size $n$ (typically, we expect $s_A, s_b = O(n^{-1/2})$), while the terms $z_{A,\delta}, z_{b,\delta}$ capture how the errors behave as a function $\delta$ (*e.g.*, they are typically of size $O(\sqrt{\ln(1/\delta)})$). In particular, $s_A, s_b$ should be independent of $\delta$ and $z_{A,\delta}, z_{b,\delta}$ should be independent of the sample size. This separation will allow us to distinguish between uniform and non-uniform versions of our oracle inequalities.

### 3.1. Minimizing the unsquared penalized loss

In this section, we present the results for the unsquared penalized loss. Choose $\|\cdot\|$ to be some norm of the $d$-dimensional Euclidean space. For $\lambda > 0$, define

$$\hat{\theta}_\lambda \in \arg\min_{\theta \in \mathbb{R}^d} \left\{ \hat{\mathcal{L}}_M(\theta) + \lambda \|\theta\| \right\}, \qquad (6)$$

where $\hat{\mathcal{L}}_M(\theta) = \|\hat{A}\theta - \hat{b}\|_M$. Our first result gives an oracle inequality for $\hat{\theta}_\lambda$ as a function of $\Delta_A$ and $\Delta_b$.

**Lemma 3.2.** *Consider $\hat{\theta}_\lambda$ as defined in* (6). *Then,*

$$\mathcal{L}_M(\hat{\theta}_\lambda) \leq \left\{1 \vee \tfrac{\Delta_A}{\lambda}\right\} \inf_{\theta \in \mathbb{R}^d} \left[\mathcal{L}_M(\theta) + (\Delta_A + \lambda)\|\theta\|\right]$$
$$+ \left\{2 \vee \left(1 + \tfrac{\Delta_A}{\lambda}\right)\right\} \Delta_b.$$

The proof, which is attractively simple and thus elegant, is given in the appendix. The result suggests that the ideal choice for $\lambda$ is $\Delta_A$. Since $\Delta_A$ is unknown, we use its upper bound to choose $\lambda$. Depending on whether we allow $\lambda$ to depend on $\delta$ or not, we get a non-uniform or uniform oracle inequality. In all cases, the rate in the oracle inequality will be fast. We start with the uniform version, non-exact version.

**Theorem 3.3.** *Let Assumption 3.1 hold and consider $\hat{\theta}_\lambda$ as defined in* (6) *where $\lambda = s_A$. Then, for any*

$0 < \delta < 1$, w.p. at least $1 - \delta$ it holds that

$$\mathcal{L}_M(\hat{\theta}_{s_A}) \leq z_{A,\delta} \cdot \inf_{\theta \in \mathbb{R}^d} \left[ \mathcal{L}_M(\theta) + s_A(1 + z_{A,\delta}) \|\theta\| \right] \\ + s_b(1 + z_{A,\delta}) z_{b,\delta} \,.$$

By allowing $\lambda$ to depend on $\delta$, we get an exact, non-uniform oracle inequality with a fast rate:

**Theorem 3.4.** *Let Assumption 3.1 hold. Fix $0 < \delta < 1$ arbitrarily and choose $\hat{\theta}_\lambda$ as defined in (6) with $\lambda = s_A z_{A,\delta}$. Then, w.p. at least $1 - \delta$ it holds that*

$$\mathcal{L}_M(\hat{\theta}_{s_A z_{A,\delta}}) \leq \inf_{\theta \in \mathbb{R}^d} \left[ \mathcal{L}_M(\theta) + 2 s_A z_{A,\delta} \|\theta\| \right] + 2 s_b z_{b,\delta} \,.$$

Note that this bound is as tight as if we had first chosen $\lambda = \Delta_A$ and then applied the stochastic assumptions to obtain a high probability (h.p.) bound.

When the linear system defined by $(A, b)$ is consistent, $\mathcal{L}_M(\theta^*) = 0$. In this case one may prefer Theorem 3.3 to Theorem 3.4. Indeed, focusing on the behavior at $\theta^*$ we get from Theorem 3.3 the bound $s_A z_{A,\delta}(1 + z_{A,\delta}) \|\theta^*\| + s_b(1 + z_{A,\delta}) z_{b,\delta}$ that holds w.p. $1 - \delta$ for any value of $\delta$, while from Theorem 3.4 we conclude the bound $2 s_A z_{A,\delta'} \|\theta^*\| + 2 s_b z_{b,\delta'}$, which however, holds only for $\delta' \geq \delta$.

### 3.2. Minimizing the squared penalized loss

A more "traditional" estimator uses the square of the empirical loss function:

$$\hat{\theta}_\rho = \arg\min_{\theta \in \mathbb{R}^d} \left\{ \hat{\mathcal{L}}_M^2(\theta) + \rho \|\theta\| \right\}, \qquad \rho > 0 \,. \qquad (7)$$

To be able to handle Lasso-like procedures, we decided to avoid squaring the norm of $\theta$. Moreover, not squaring this term is convenient for the proof techniques we used. The extension of our results for other types of penalties, in particular $\|\theta\|^2$, is left for future work.

Unlike the previous case where the loss function and the norm were both unsquared, in this case the selection of the regularization parameter $\rho$ will be more involved. In practice, one often uses a hold-out estimate to choose the best value of $\rho$ amongst a finite number of candidates on an exponential grid. Here, we propose a procedure that avoids splitting the data, but uses the unsquared penalized loss with the same data. The new procedure is defined as follows. For some $\lambda, c > 0$ to be chosen later, let

$$\hat{\rho}(\lambda, c) \in \arg\min_{\rho \in \Lambda(\lambda, c)} \left\{ \hat{\mathcal{L}}_M(\hat{\theta}_\rho) + \lambda \|\hat{\theta}_\rho\| \right\}, \qquad (8)$$

where $\Lambda(\lambda, c) \doteq \{ 2^k \cdot 2c\lambda : k \in \mathbb{N} \}$ and define

$$\tilde{\theta}_{\lambda,c} \doteq \hat{\theta}_{\hat{\rho}(\lambda,c)} \,. \qquad (9)$$

We now have two parameters that need tuning. However, as we will see, the tuning of these parameters is very similar to what we have seen in the previous section. The reason for this is that $\Lambda$ is rich enough to contain a value $\rho$ that makes $\hat{\mathcal{L}}_M(\hat{\theta}_\rho) + \rho \|\hat{\theta}_\rho\|$ comparable to (not much larger than) $\hat{\mathcal{L}}_M(\theta) + \lambda \|\theta\|$ no matter what $\theta$ one selects. This is in fact the key to the proof of the following lemma, which gives a deterministic oracle inequality for $\tilde{\theta}_{\lambda,c}$:

**Lemma 3.5.** *Let $\tilde{\theta}_{\lambda,c}$ be as in (9). Then,*

$$\mathcal{L}_M(\tilde{\theta}_{\lambda,c}) \leq \left\{ 1 \vee \tfrac{\Delta_A}{\lambda} \right\} \inf_{\theta \in \mathbb{R}^d} \left[ \mathcal{L}_M(\theta) + (\Delta_A + 2\lambda) \|\theta\| \right] \\ + \left\{ 2 \vee \left(1 + \tfrac{\Delta_A}{\lambda}\right) \right\} \Delta_b + \left\{ 1 \vee \tfrac{\Delta_A}{\lambda} \right\} c.$$

With the (unattainable) choice $\lambda = \Delta_A, c = \Delta_b$ we get

$$\mathcal{L}_M(\tilde{\theta}_{\lambda,c}) \leq \inf_{\theta \in \mathbb{R}^d} \left[ \mathcal{L}_M(\theta) + 3\Delta_A \|\theta\| \right] + 3\Delta_b.$$

These choices are impractical but, as it happened with in the previous section, we can obtain uniform non-exact or non-uniform exact oracle inequalities with fast rates. The non-exact uniform oracle inequality is formalized as follows:

**Theorem 3.6.** *Let Assumption 3.1 hold and choose $\tilde{\theta}_{\lambda,c}$ be as in (9) with $\lambda = s_A$ and $c = s_b$. Then, for any $0 < \delta < 1$ w.p. at least $1 - \delta$ it holds that*

$$\mathcal{L}_M(\tilde{\theta}_{\lambda,c}) \leq \{1 \vee z_{A,\delta}\} \inf_{\theta \in \mathbb{R}^d} \left[ \mathcal{L}_M(\theta) + s_A(z_{A,\delta} + 2) \|\theta\| \right] \\ + \{2 \vee (1 + z_{A,\delta})\} s_b z_{b,\delta} + \{1 \vee z_{A,\delta}\} s_b \,.$$

The next theorem gives a non-uniform, exact oracle inequality with fast rates.

**Theorem 3.7.** *Let Assumption 3.1 hold. Fix $0 < \delta < 1$ and choose $\tilde{\theta}_{\lambda,c}$ be as in (9) with $\lambda = s_A z_{A,\delta}$ and $c = s_b z_{b,\delta}$. Then, w.p. at least $1 - \delta$ it holds that*

$$\mathcal{L}_M(\tilde{\theta}_{\lambda,c}) \leq \inf_{\theta \in \mathbb{R}^d} \left[ \mathcal{L}_M(\theta) + 3 s_A z_{A,\delta} \|\theta\| \right] + 3 s_b z_{b,\delta} \,.$$

The relative merits of the uniform and non-uniform oracle inequalities are unchanged compared to what we have seen in the previous section.

## 4. Value-estimation in Markov Reward Processes: Results

Let us now return to value-estimation in Markov Reward Processes. We consider the projected Bellman error objective, $\mathcal{L}_M(\theta) = \|A\theta - b\|_M$, where $M = C^{-1}$ (for the definitions see Section 2). Assume that $\Delta_A$, $\Delta_b$ are concentrated as in Assumption 3.1, with known

bounds. This can be arranged for example if the features $\phi_i(X_t)$ and rewards $R_{t+1}$ are a.s. bounded, and if we assume appropriate mixing, such as exponential $\beta$-mixing (Yu, 1994), or when the Markov chain $(X_t)_{t\geq 0}$ forgets its past sufficiently rapidly (Samson, 2000). Note that in these cases $(\hat{A}, \hat{b})$ gets concentrated around $(A, b)$ at the usual parametric rate, i.e., $s_A, s_b = O(\sqrt{1/n})$ and $z_{A,\delta}, z_{b,\delta} = O(\sqrt{\ln(1/\delta)})$.

For simplicity, assume first that $C$ is given and consider the *on-policy case*. As mentioned previously, in this case the system $A\theta = b$ is guaranteed to have a solution and therefore $\mathcal{L}_M(\theta^*) = 0$. Consider the estimator that minimizes the unsquared penalized loss. Then, Theorem 3.3 shows a uniform fast rate when using $\lambda = s_A$:

$$\mathcal{L}_M(\hat{\theta}_{s_A}) \leq (1 + z_{A,\delta})\Big[s_A z_{A,\delta}\|\theta^*\| + s_b z_{b,\delta}\Big].$$

We get a similar inequality for the squared penalized loss using the result Theorem 3.6 with a slightly larger bound.

In the *off-policy case*, the linear system $A\theta = b$ may not have a solution. When it does, the previous bound applies. However, when this linear system does not have a solution, to get an exact oracle inequality we are forced to choose $\lambda$ (in the case of minimizing the unsquared penalized loss) based on $\delta$. In particular, with the choice $\lambda = s_A z_{A,\delta}$, Theorem 3.4 gives

$$\mathcal{L}_M(\hat{\theta}_{s_A z_{A,\delta}}) \leq \inf_{\theta \in \mathbb{R}^d} \Big[\mathcal{L}_M(\theta) + 2 s_A z_{A,\delta}\|\theta\|\Big] + 2 s_b z_{b,\delta}. \tag{10}$$

Again, this inequality gives fast, $O(\sqrt{1/n})$ rates when $s_A, s_b = O(\sqrt{1/n})$. Similar results hold for the procedure defined for the squared penalized loss where the bound is given by the inequality of Theorem 3.7.

When $C$ is unknown, one may resort replacing it by $M \succ 0$. Then, a non-exact oracle inequality can be derived using $\|x\|_P^2 \leq \nu_{\max}(Q^{-1/2}PQ^{-1/2})\|x\|_Q^2$. (For a matrix $S$, we denote by $\nu_{\max}(S), \nu_{\max}(S)$ its largest and smallest singular values, respectively.) Consider first the unsquared penalized loss. In this case, $\|A\theta - b\|_{C^{-1}} \leq \nu_{\max}^{1/2}(M^{-1/2}C^{-1}M^{-1/2})\|A\theta - b\|_M$. Assume that for an estimator $\hat{\theta}$ it holds that $\|A\hat{\theta} - b\|_M \leq \inf_\theta \Big[\|A\theta - b\|_M + c_{\hat{A},\hat{b}}(\theta)\Big]$. Then, from $\|A\theta - b\|_M \leq \nu_{\max}^{1/2}(C^{1/2}MC^{1/2})\|A\theta - b\|_{C^{-1}}$ we get

$$\|A\hat{\theta} - b\|_{C^{-1}} \leq \inf_\theta \Big[\kappa^{1/2}\|A\theta - b\|_M + \tau^{-1/2} c_{\hat{A},\hat{b}}(\theta)\Big].$$

where $\kappa = \nu_{\max}(C^{1/2}MC^{1/2})/\nu_{\min}(M^{1/2}CM^{1/2})$ is the "conditioning number" of $M^{1/2}CM^{1/2}$ and $\tau = \nu_{\min}(M^{1/2}CM^{1/2})$. In the on-policy case, for example, this gives bounds of the form

$$\mathcal{L}_M(\hat{\theta}_{s_A}) \leq \tau^{-1/2}(1 + z_{A,\delta})\Big[s_A z_{A,\delta}\|\theta^*\| + s_b z_{b,\delta}\Big].$$

The bound for the off-policy case derived from (10) takes the form

$$\mathcal{L}_M(\hat{\theta}_{s_A z_{A,\delta}}) \leq$$
$$\inf_{\theta \in \mathbb{R}^d} \Big[\kappa^{1/2}\mathcal{L}_M(\theta) + 2\tau^{-1/2} s_A z_{A,\delta}\|\theta\|\Big] + 2\tau^{-1/2} s_b z_{b,\delta}.$$

Similar inequalities can be derived for our procedures that minimize the squared penalized loss.

Finally, let us discuss the dependence of our bounds on the choice of the basis functions. This dependence comes through Assumption 3.1. As an example, assume that $\phi_i : \mathcal{X} \to [-1, 1]$ and $\|\cdot\| = \|\cdot\|_p$ with $1 \leq p \leq 2$. In this case, the bound on $\Delta_A$ is expected to scale linearly with $d$, while $\Delta_b$ is expected to scale linearly with $\sqrt{d}$. To see why $\Delta_A$ is expected to scale linearly with $d$ note that $\Delta_A \leq \|M^{1/2}(\hat{A} - A)\|_{2,2} = \|M^{1/2}(\hat{A} - A)\|_F$, where $\|\cdot\|_F$ denotes the Frobenius norm. Now, the Frobenius norm is the norm underlying the Hilbert-space of square matrices with the inner product $\langle P, Q \rangle = \text{trace}(P^\top Q)$ and thus an application of any concentration inequality for Hilbert-space valued random variables (e.g., (Steinwart and Christmann, 2008)) gives a bound that scales with the "range" of $N = \|M^{1/2}(\phi(X_t) - \gamma\phi(\tilde{X}_{t+1}))\phi(X_t)^\top\|_F$. Using the rotation property of trace, we get that $N = \|\phi(X_t) - \gamma\phi(\tilde{X}_{t+1})\|_M \|\phi(X_t)\|$. The first term can be bounded using the triangle inequality as a function of $\|\phi(X_t)\|_M$ and $\|\phi(\tilde{X}_{t+1})\|_M$. Assuming (e.g.,) that $M$ is the identity matrix, we get that both $\|\phi(X_t)\|_M = \|\phi(X_t)\|$ and $\|\phi(\tilde{X}_{t+1})\|$ are of size $O(\sqrt{d})$. Hence, their product scales linearly with $d$.

The above bound on $\Delta_A$ is naive; we believe using $\Delta_A \leq \nu_{\max}(\hat{A} - A)$ may yield a tighter dependency on $d$. E.g., for $d \times d$-matrices with i.i.d standard normal entries, the maximum eigenvalue is $O(\sqrt{d})$ (Vershynin, 2010). Furthermore, note that if the basis functions are correlated, or if they are sparse, the dimension will not necessarily appear linearly in the bound either. For a discussion of when to expect a milder dependence of the norm of $\phi$ on $d$, the interested reader may consult the paper by Maillard and Munos (2009).

### 4.1. Related work

Antos et al. (2008) proved a uniform high-probability inequality both for the on-policy and the *off-policy cases* for LSTD. Their bound takes the form $\mathcal{L}_M(\hat{\theta}) - \mathcal{L}_M(\theta^*) = O\left(d\ln(d)\left(\frac{1}{n}\right)^{\frac{1}{4}}\right)$, which is a slower rate than the rate we are able to obtain. Further, with our bounding method the $\ln d$ factor can be removed from this bound.

There are more results available for the on-policy case. As mentioned earlier, in this case the system $A\theta = b$ is consistent and thus our bound, under appropriate mixing conditions, takes the form

$$\mathcal{L}_M(\hat{\theta}) = O\left(L\sqrt{\frac{d}{\tau n}}(1+R)\right),$$

where $\tau \doteq \nu_{\min}(M^{\frac{1}{2}}CM^{\frac{1}{2}})$, $L$ is the worst-case norm of features in the dual norm ($L \doteq \sup_{x \in \mathcal{X}} \|\phi(x)\|_*$; as discussed previously, $L$ may be $O(\sqrt{d})$) and $R$ is a worst-case bound on the norm of the parameter vector (i.e., $\|\theta^*\| \leq R$). In the next two results, the norm $\|\cdot\|$ is the 2-norm. Lazaric et al. (2010) for their (unregularized) path-wise LSTD method obtain

$$\mathcal{L}_M(\hat{\theta}) = O\left(L\sqrt{\frac{d \log d}{n\tau}}(1+R)\right)$$

(cf. Theorem 3 in their work). Although this is a fast rate, it also shares the undesirable dependence on $\frac{1}{\tau}$. Non-uniform, slow rates can be extracted from the paper by Ghavamzadeh et al. (2010) for LSTD with random projections. The result with our notation would look like (cf. Theorem 2)

$$\mathcal{L}_M(\hat{\theta}) = O\left(L^2\sqrt{\frac{\log d}{\tau}}\left(\frac{1}{n}\right)^{\frac{1}{4}}R + \frac{LR}{\sqrt{n}}\right).$$

More recently, for the so-called Lasso-TD method, Ghavamzadeh et al. (2011) showed non-uniform $O\left(\left(\frac{1}{n}\right)^{\frac{1}{4}}\right)$-rates, but only for the so-called in-sample error, i.e., the empirical norm at the states used by the algorithm. These rates depend on the $\ell^1$-norm of $\theta^*$ and have no dependence on the minimum eigenvalue, but they are slow in $n$. At the expense of additional assumptions on the Gram matrix $\hat{C}$ (a sample estimate of $C$), they have also derived fast rates.

## 5. Conclusion and future work

We have shown performance bounds for two estimators in linear inverse problems. Each of these minimizes one of $\mathcal{L}_M(\theta)$ and $\mathcal{L}_M^2(\theta)$, plus a penalty $\lambda\|\theta\|$. The penalty weight $\lambda$ can be chosen a priori without the need for a separate validation data set, and the bounds were presented in a general form that apply to many different instances of statistical linear inverse problems, requiring only that $\Delta_A$ and $\Delta_b$ concentrate around zero. Our split analysis, into a deterministic step and a stochastic step, allows us to decouple the behavior of $\Delta_A, \Delta_b$ from that of the estimators.

We have recovered $\ell^1$-penalized variations of LSTD (Bradtke and Barto, 1996) for value function estimation in MRPs. We have shown fast, uniform rates, which, in the on-policy case, are exact and competitive with those existing in the literature. In the off-policy case, the rates are non-exact, and the non-uniform bound is also competitive with existing results.

Finally, we would like to point out interesting ways to further develop our work.

$\ell^1$-**penalties.** The choice when the norm used in the penalty is the $\ell^1$-norm has been extensively studied in the supervised learning literature (see, e.g., (Bickel et al., 2009; Koltchinskii, 2011; Bühlmann and Geer, 2011) and the references therein), as well as in the reinforcement learning setting (Kolter and Ng, 2009; Ghavamzadeh et al., 2010; 2011; Maillard, 2011), mainly because it allows for non-trivial performance bounds even when the dimension $d$ of the parameter vector is comparable to the sample size $n$ (or even larger than $n$) provided that the true parameter vector is sparse (i.e., there are many zeroes in it). In this paper we decided not to specialize to this case but rather to focus on the problem of proving fast, exact and (possibly uniform) oracle inequalities. Our results, when applied to the case of an $\ell^1$-penalty show that in a way adding an $\ell^1$-penalty does not hurt performance (as we expect that the oracle inequalities with the said properties should hold for a decent method) even if the conditions ideal for the $\ell^1$-penalty do not hold. We do not know of performance bounds (ours included) for $\ell^1$-penalized estimation have *all* of the characteristics we are after in a bound (*viz.* bounds that are exact, fast and uniform).

**Linear regression.** Our results are also worth investigating in the context of linear regression. It is easy to cast regression as a statistical linear estimation problem whose underlying system is always consistent. If we use $\|\cdot\|$ as the $\ell^1$-norm, we recover procedures similar to the square-root Lasso (Belloni et al., 2010) and the Lasso (Tibshirani, 1996) for the estimators studied in Sections 3.1 and 3.2, respectively. We believe that confronting the bounds that can be derived from our results with bounds for linear regression in the literature can be very instructive.

**Connection to Inverse Problems.** The theory of Inverse Problems is very pertinent to this work, and it is important to study our results under the light of those shown in Chapter 2 of Kirsch (2011); Alquier et al. (2011). The existing knowledge of inverse prob-lems may help us better understand which choices of $\|\cdot\|$ allow $\Delta_A$ to concentrate around zero, and how fast this concentration occurs. The idea of having learning problems as inverse problems is not new; Rosasco (2006); Vito et al. (2006) study regression in Hilbert spaces as an inverse problem.

## Acknowledgements

This work was supported by AITF and NSERC.